\documentclass[conference]{IEEEtran}
\IEEEoverridecommandlockouts
\usepackage{cite}
\usepackage{amsmath,amssymb,amsfonts}
\usepackage{algorithmic}
\usepackage{graphicx}
\usepackage{textcomp}
\usepackage{xcolor}
\usepackage{hyperref}
\usepackage{multirow}
\def\BibTeX{{\rm B\kern-.05em{\sc i\kern-.025em b}\kern-.08em
    T\kern-.1667em\lower.7ex\hbox{E}\kern-.125emX}}
\begin{document}

\title{Object Detection with Spiking Neural Networks \\ on Automotive Event Data
\thanks{This material is based upon work supported by the French technological research agency (ANRT) through a CIFRE thesis in collaboration with Renault.}
}

\author{\IEEEauthorblockN{Loïc Cordone}
\IEEEauthorblockA{\textit{Renault}\\
\textit{University Côte d'Azur} \\
\textit{LEAT / CNRS UMR 7248}\\
loic.cordone@renault.com}
\and
\IEEEauthorblockN{Benoît Miramond}
\IEEEauthorblockA{\textit{University Côte d'Azur} \\
\textit{LEAT / CNRS UMR 7248}\\
Sophia Antipolis, France \\
benoit.miramond@univ-cotedazur.fr}
\and
\IEEEauthorblockN{Philippe Thierion}
\IEEEauthorblockA{\textit{Software Factory}\\\textit{Renault}\\
Sophia Antipolis, France \\
philippe.thierion@renault.com}
}


\maketitle

\begin{abstract}
Automotive embedded algorithms have very high constraints in terms of latency, accuracy and power consumption. In this work, we propose to train spiking neural networks (SNNs) directly on data coming from event cameras to design fast and efficient automotive embedded applications. Indeed, SNNs are more biologically realistic neural networks where neurons communicate using discrete and asynchronous spikes, a naturally energy-efficient and hardware friendly operating mode. Event data, which are binary and sparse in space and time, are therefore the ideal input for spiking neural networks. But to date, their performance was insufficient for automotive real-world problems, such as detecting complex objects in an uncontrolled environment. To address this issue, we took advantage of the latest advancements in matter of spike backpropagation - surrogate gradient learning, parametric LIF, SpikingJelly framework - and of our new \textit{voxel cube} event encoding to train 4 different SNNs based on popular deep learning networks: SqueezeNet, VGG, MobileNet, and DenseNet. As a result, we managed to increase the size and the complexity of SNNs usually considered in the literature. In this paper, we conducted experiments on two automotive event datasets, establishing new state-of-the-art classification results for spiking neural networks. Based on these results, we combined our SNNs with SSD to propose the first spiking neural networks capable of performing object detection on the complex GEN1 Automotive Detection event dataset. 

\end{abstract}


\section{Introduction}
Event cameras have several interesting features for embedded automotive applications: very low latency, high dynamic range and low power consumption. They are composed of independent photoreceptor pixels that detect a change in brightness. They output binary events, a type of information that contains the position, the precise time and polarity of every brightness change. But since the data they produce is asynchronous and sparse by nature, it is difficult to use classical neural networks that were designed for processing frames.

In spiking neural networks, the information is transmitted between neurons by discrete binary spikes, making the whole network asynchronous and sparse by nature. They are therefore particularly adapted to the processing of event data. Due to the nature of their operations, they are also hardware friendly: previous works showed that on specialized hardware, spiking neural networks consume 50\% less energy than traditional neural networks while maintaining the same accuracy \cite{lyesijcnn}, and the recent release of new neuromorphic hardware such as Intel Loihi 2 \cite{loihi2} could further improve these results.

Using the latest advancements in matter of spike backpropagation, we train state-of-the-art spiking neural networks to process real-world automotive event data and obtained good performance on classification and object detection tasks. 

The main contributions of this work can be summarized as follows:
\begin{enumerate}
    \item We present a novel approach to encode event data called \textit{voxel cube} that preserves their binarity and temporal information while keeping a low number of timesteps.
    \item We propose a new challenging dataset for classification on automotive event data: GEN1 Automotive Classification, generated using the Prophesee object detection dataset of the same name.
    \item We train four different spiking neural networks for classification tasks based on popular neural network architectures (SqueezeNet, VGG, MobileNet, DenseNet) and evaluate them on two automotive event datasets, setting new state-of-the-art results for spiking neural networks. 
    \item We present spiking neural networks for object detection composed of a spiking backbone and SSD bounding box regression heads that achieve qualitative results on the real-world GEN1 Automotive Detection event dataset.
    To the best of our knowledge, it constitutes the first spiking neural networks capable of doing object detection on real-world event dataset. 
\end{enumerate}

Our code is available upon request and will be available online in the future.

\section{Related Work}
\subsection{Learning on event data}
With its microsecond temporal resolution, event data cannot be directly processed by modern deep learning models. Over the years, several preprocessing methods have emerged to convert events into a dense representation that can be used in Deep Neural Networks (DNNs).

The simplest preprocessing method is to accumulate the events pixel-wise over a given period of time, creating an histogram \cite{steering2018}. The result can be viewed as an event frame, with no time dimension and two output channels containing the event count for each polarity. Time surfaces \cite{hots}, where the timestamp of the last received event is stored in each pixel, represent an alternative to histograms that better preserve the temporal information inside an event frame. A recent preprocessing method called "event cube" \cite{eventcube} has been proposed to combine the simplicity of histograms with the temporal information of time surfaces. In event cubes, a given period of time is split in $n$ micro time bins, and each pixel stores not the event count but the event temporal distance from the center of the neighboring micro time bins. The temporal information is thus contained in the channel dimension, of size $2 \times n$.

The simplest way to retain the temporal information is to represent events as a voxel grid \cite{voxel}, where each voxel represents a pixel and a time interval. Although it requires more memory and more computation, this is the preferred representation for spiking neural networks as both the data and the network operate on a fixed number of timesteps. The accumulation of events on the time interval is usually a sum, as in \cite{plif}, but a binary accumulation as in \cite{loicijcnn} has the advantage of interpreting events as binary spikes, and therefore benefit from the spiking neural networks energy gains on specialized hardware. 

This paper introduces a novel representation called "voxel cube" that combines binary voxel grids with event cubes to obtain a binary representation with a small number of timesteps that maintains high temporal information.

\subsection{Spiking Neural Networks}

Spiking neural networks more closely mimic biological neural networks by incorporating the concept of time into their operating model. The neurons communicate using binary spikes following a spiking neuron model e.g. the Leaky Integrate-and-Fire (LIF) neuron \cite{lif}, that is widely used due to the simplicity of its operations. Neurons are connected by scalar weights, modeling the synapses. As in DNNs, it is possible to learn these weights, but since spikes are discrete and thus non-differentiable, it is not possible to use the popular backpropagation learning algorithm on SNNs. For static data, SNNs can be created by converting to the spike domain a trained DNN, using rate coding for example. However, the results are inevitably inferior, making the direct learning of SNNs more relevant especially for temporal data. To overcome the lack of a backpropagation algorithm on spikes, several learning rules have been proposed over the years. 


Spike-timing-dependent plasticity (STDP) \cite{stdp} is a bio-plausible unsupervised learning rule where the weight connecting two neurons is modified according to the delay between the firing of the presynaptic and postsynaptic neurons. For a long time, results obtained with STDP were not competitive with supervised learning even on simple tasks, but recent progress \cite{stdpgood} could change the picture. However, STDP still has not proved its effectiveness on real-word complex tasks such as object detection.

Recently, supervised learning rules based on backpropagation have enabled the training of SNNs with excellent results. Reference \cite{slayer} presented SLAYER, an error backpropagation method for SNNs capable of learning both synaptic weights and axonal delays, allowing them to tackle bigger datasets with deeper networks. Another approach introduced in \cite{neftci} is to train SNNs using surrogate gradient learning: during the forward pass, an Heaviside step function generates spikes and during the backward pass, the gradient of this non-differentiable function is approximated using a surrogate gradient, for example the gradient of a sigmoid function. This work also demonstrated that SNNs are equivalent to RNNs, enabling their learning with backpropagation through time in popular deep learning frameworks.

Using this surrogate gradient learning rule, multiple Pytorch-based frameworks for training SNNs have emerged, such as Nengo \cite{nengo} or SpikingJelly \cite{spikingjelly}. Thanks to their automatic differentiation and strong GPU acceleration, these new frameworks have led to the training of deeper spiking neural networks achieving state-of-the-art results on classification problems \cite{plif}, \cite{wei_neurips}. It also becomes possible to make the neuron parameters learnable. For example, reference \cite{plif} introduced the Parametric LIF (PLIF) neuron model, where the learnable time constants make the network less sensitive to initial values and can speed up learning.

\subsection{Object Detection on event data}

The task of object detection is to determine where objects are located in a given image (object localization) and which category each object belongs to (object recognition). DNNs have brought significant performance gains over handcrafted methods by learning semantic, high-level and deeper features. For embedded real-time applications, one-step object detection frameworks such as SingleShot MultiBox Detector (SSD) \cite{ssd} have emerged. They map image pixels to bounding box coordinates and class probabilities, modeling object detection as a regression or classification task.

Object detection on event data, which is sparse and contains high temporal information, is still a challenge. The simplest approach is to integrate the events as a single frame, and to train a deep neural network without temporal information. This is used for example in the YOLE network\cite{yole}. With an event preprocessing method that retains temporal information, it is possible to design more complex object detection networks. 
In \cite{matrixlstm} authors propose Matrix-LSTM a network based on voxel grids representations and composed of a matrix of LSTM cells that sequentially processes event features. These features are then fed to a pretrained object detection model. A more ambitious approach is to train an unique neural network directly on voxel grids, using the temporal information to output the bounding boxes and the classes. Such an approach is presented in \cite{1mpx}: Recurrent Event-camera Detector (RED) is a recurrent neural network composed of feed-forward convolutional layers followed by ConvLSTM. While the convolutional layers extract low-level features from events, the ConvLSTM layers extract high-level spatio-temporal patterns thanks to their memory. The output is then fed to SSD bounding box regression heads.

This demonstrates that learning to detect objects from events is possible, but it still requires deep neural networks that do not take full advantage of the properties of event-data and that would be difficult to embed in power constrained environment. Hybrid-SNN \cite{hybridsnn} proposes a partial solution by presenting an hybrid neural network composed of a SNN backbone for efficient event-based feature extraction, and an ANN head to solve object detection tasks. To the best of our knowledge, the work presented in our paper is the first complete spiking neural network capable of doing object detection.

\section{Method}

\subsection{Voxel cubes}

Most event sensors output events with  microsecond temporal resolution, but in order to use them in modern deep learning models, we need to convert them into a dense representation. In our case, we accumulate events over time windows of $\Delta t$ seconds. The resulting representation is called a voxel grid, where each voxel represent a pixel and a time interval. Indeed, a sample lasting $d$ seconds is now divided in $\frac{d}{\Delta t} = T$ timesteps. Usually, the events are stored in the form of a 4D $CTHW$ tensor, with $C$ the number of channels, $T$ the number of timesteps, $H$ and $W$ the height and width of the data. 

We can then use spiking neural networks directly on event data, they will therefore operate on $T$ timesteps. But in order to keep the high temporal resolution of event data, we need to have a large number of timesteps $T$, which increases linearly the number of computations of the SNN and thus the inference time and the energy consumed. 

\begin{figure}[tb]
\centerline{\includegraphics[scale=0.15]{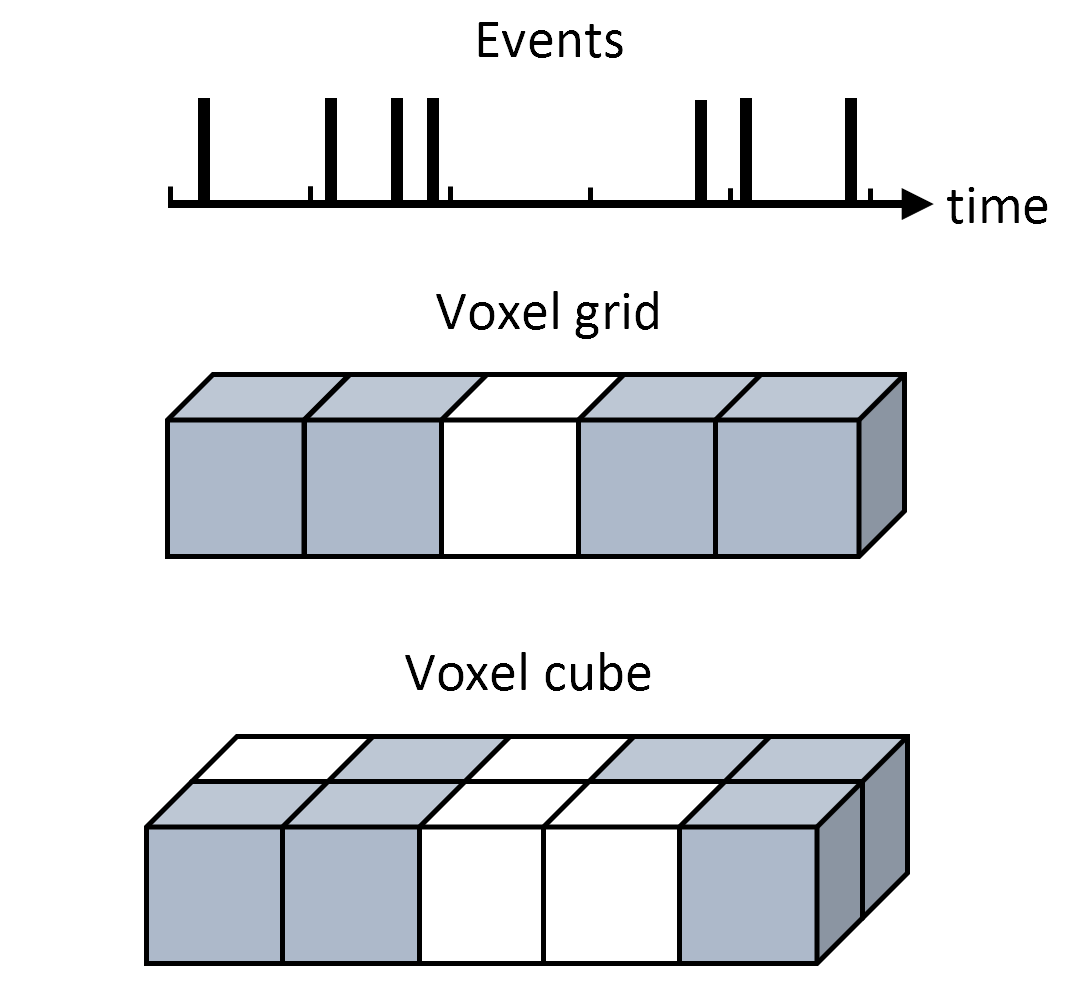}}
\caption{Voxel cube encoding. For a given number of timesteps, voxel cubes better preserve the temporal information of events by exploiting the channel dimension. Here, the voxel cube illustrated uses 2 micro time bins.}
\label{fig:voxelcube}
\end{figure}

Inspired by event cubes \cite{eventcube}, we propose a novel event preprocessing called \textit{voxel cubes}. In voxel cubes, each time window $\Delta t$ is subdivided in $n$ micro time bins lasting therefore $\frac{\Delta t}{n}$ seconds. Events belonging to a micro time bin will be stored in the channels dimension, providing finer temporal information to the first layer of the network. The number of channels $C$ is therefore equals to $2 \times n$, each polarity being stored in $n$ channels. Contrary to event cubes, each event contributes only to the time bin where it falls into, and the accumulation of multiple events in the same micro time bin is binary. This loss of information is justified by the need to keep binary inputs, in order to leverage the energy efficiency of spiking neural networks running on specialized hardware. An illustration of this encoding is proposed in Fig.~\ref{fig:voxelcube}.

By transferring temporal information to the channel dimension, voxel cubes allow us to reduce the number of timesteps $T$ without loosing temporal precision compared to voxel grids, as we show in section \ref{tbin}.

\subsection{Spiking Neural Networks models}

Inspired by popular deep learning convolutional neural networks (CNNs), we designed 4 different spiking neural networks using only strided convolutions, max pooling, batch normalization and PLIF neurons. Convolutions and max pooling have been used extensively with SNNs \cite{loicijcnn} \cite{pool}. Since batch normalization layers can be merged with a preceding or subsequent convolution layer at inference, it is possible to use them to train SNNs as long as they are placed before the PLIF neurons. We explore their importance in section \ref{bn}. 

In CNNs, the final layers used for classification need to be adapted to be compatible with spikes. To this end, we propose a spiking classifier simply composed of a layer of batch normalization, a $1\times 1$ convolution outputting $num\_classes$ channels and PLIF neurons. By using a 1D convolution, this classifier is able to process feature maps of any size without requiring e.g. a layer of average pooling, which would be incompatible with spikes operations. The final predictions are obtained by summing all output spikes first in the spatial dimension, then in the time dimension. 

We propose spiking variants of VGG, SqueezeNet, MobileNet and DenseNet by replacing their ReLU activation functions by PLIF neurons. All spiking neural networks use the spiking classifier described above in lieu of their own classifier. \\



\subsubsection{Spiking VGG} 
Introduced in \cite{vgg}, VGG is a convolutional neural network composed of up to 19 convolutional layers followed by 3 fully-connected layers. Our Spiking VGG replaced the final classifier but kept the same architecture, with the addition of batch normalization before each spiking convolutional layer. \\

\subsubsection{Spiking SqueezeNet}
SqueezeNet \cite{squeezenet} is a small CNN that uses Fire modules: squeeze layers ($1\times 1$ convolutions) before expand layers composed of a mix of $1\times 1$ and $3\times 3$ convolutions. This results in a low number of parameters, an appealing property for embedded applications. We replaced all convolutional layers by their spiking equivalent to obtain our Spiking SqueezeNets, along the addition of batch normalization. \\

\subsubsection{Spiking MobileNet}
MobileNet \cite{mobilenet} is a model designed to be used in mobile applications that use depthwise separable convolutions, requiring less parameters and computations than normal convolutions. For our Spiking MobileNet, we dropped the activation function between the depthwise and pointwise convolutions, and moved all batch normalization layers before the convolutional layers. Removing the activation function makes our network non-spike as the inputs of the pointwise convolution are not spikes anymore. However, it can be shown that a depthwise separable convolution is equivalent to a normal convolution with specific weights. Thus, we used depthwise separable convolutions in the training of our spiking MobileNets as it provided better results (see section \ref{dw}), and we return to a full-compatible SNN at inference by replacing them by their equivalent convolutions. Our Spiking MobileNet contains only one of the five identical depthwise separable layers near the end of the network. We have varied the number of filters of the first layer to obtain networks of different sizes. \\

\subsubsection{Spiking DenseNet}
To promote gradient propagation, ResNets use element-wise addition, an operation difficult to operate in the spike domain. Reference \cite{wei_neurips} proposed Spiking ResNets with different residual connections, but the one based on an AND accumulation - in our opinion the only one compatible with an implementation on specialized hardware - produces unsatisfactory results. DenseNet \cite{densenet} is an architecture that promotes gradient propagation by using channel-wise concatenations, which is an operation preserving the spike representation. We replaced the ReLU activations by PLIF neurons to obtain our Spiking DenseNet. We have varied the depth and growth rate to obtain different versions of Spiking DenseNet.


\subsection{Object Detection with Spiking Neural Networks}

\begin{figure*}[tb]
\centerline{\includegraphics[scale=0.30]{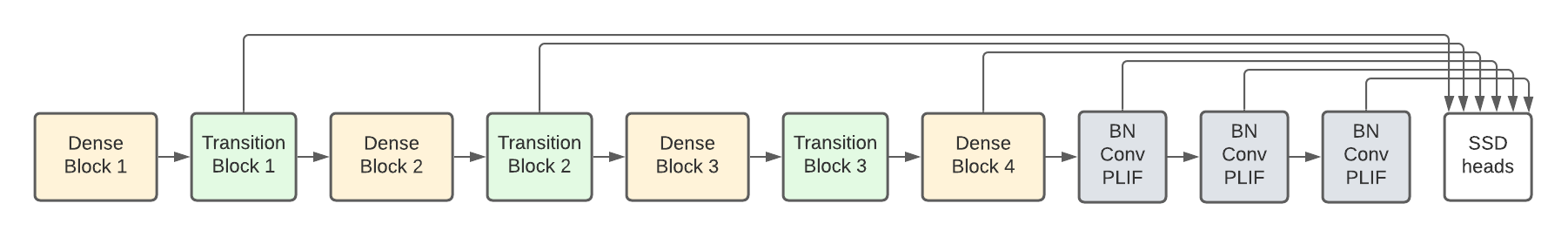}}
\caption{Our spiking DenseNet + SSD architecture for object detection.}
\label{fig:od}
\end{figure*}

The SSD object detection framework \cite{ssd} consists of a backbone and multiple predictor heads. The heads take as inputs feature maps generated by the backbone at different scales to predict bounding boxes and their associated classes. To obtain a complete spiking neural network capable of doing object detection, we replaced the CNN backbone by the SNN backbones designed for classification, and we used spiking convolutions instead of normal convolutions in the extra layers. Therefore, the feature maps fed to the SSD heads are spikes, and since the heads consist of only one convolution the whole network is indeed a SNN.

The spiking neural network operates on $T$ timesteps, therefore the final bounding boxes and classes predicted by the network are obtained by doing a sum over the $T$ timesteps. The output of the network still requires a post-processing to filter predictions, but we assume that this step would be done on conventional hardware outside the spiking neural network scope.

One-shot object detectors such as SSD struggles with the class imbalance problem due to the overwhelming number of predictions classified as background. Reference \cite{focalloss} introduced focal loss, a modulation term applied to the cross-entropy loss function that tremendously helps the learning of one-shot object detectors. Thus, we trained our spiking neural network with focal loss, as the hard negative mining originally used by SSD did not achieved satisfying performances.

As in the original SSD architecture, we used three extra blocks of convolutions to generate smaller feature maps after our spiking backbone. Each block consists of a $1 \times 1$ spiking convolution to reduce the number of channels, followed by a $3 \times 3$ spiking convolution with a stride of 2. Once again, we used batch normalization layers before each convolution. 

The anchors used by our network were generated with a minimum ratio of 0.5 and a maximum ratio of 0.8, accounting for the smaller objects present in the object detection dataset we studied.

The architecture of our spiking DenseNet + SSD is shown in Fig.~\ref{fig:od}. Increasingly smaller feature maps generated by the DenseNet backbone are fed to the SSD heads, and the 3 extra blocks further reduce features maps to a final size of $2\times1$. Similar architectures are used for our spiking MobileNet + SSD and VGG + SSD.

\section{Experiments}

\subsection{Datasets}

We evaluated our spiking neural networks on two automotive classification datasets: Prophesee NCARS and a new event dataset we called Prophesee GEN1 Automotive Classification, generated from the object detection dataset Prophesee GEN1. We then evaluated our object detection spiking networks on this specific dataset, the Prophesee GEN1 Automotive Detection dataset.

\subsubsection{Prophesee NCARS dataset} 
The Prophesee NCARS dataset \cite{ncars} is a classification composed of 24k samples of length 100ms captured with a Prophesee GEN1 event camera mounted behind the windshield of a moving car. The samples represent either a car or background. Samples have variable size as they are cropped from recordings of resolution $304 \times 240$ pixels.

\subsubsection{Prophesee GEN1 Detection dataset} 
Composed of 39 hours of recording, the Prophesee GEN1 Automotive Detection dataset \cite{gen1} is the largest event-based dataset to date. Recorded with a Prophesee GEN1 sensor mounted on a car dashboard, it contains over 255k manually annotated bounding boxes of two classes: cars and pedestrians.

\subsubsection{Prophesee GEN1 Classification dataset} 
We generated a classification dataset from the Prophesee GEN1 Detection dataset by cropping each bounding box (car or pedestrian) form individual samples. As it is the case for NCARS, each sample represents 100ms of events preceding the annotated bounding box. The main difference between this dataset and NCARS lies in the presence of a pedestrian class. Indeed, we believe that the features learned by a network are more relevant if it is trained to classify two different classes rather than one class vs. background. 

To avoid imbalance in the number of samples for each class, we rebalanced the training set: we undersampled the cars and oversampled the pedestrians by doing horizontal flip data augmentation. The code used to generate this classification dataset will be available online.

\subsection{Implementation details}
\subsubsection{Voxel Cubes \label{voxel}}
For classification, we used samples of 100ms encoded as binary voxel cubes. For object detection, we chose to train our network only on the 100ms preceding annotated bounding boxes. Thus, our SNN make predictions based solely on these 100ms, the potentials being reset after each sample. 
For both tasks, the samples were encoded as binary voxel cubes of 5 timesteps and 2 micro time bins, as it represented the best compromise between performance and number of operations. 

\subsubsection{Hyperparameters}
All classification models were trained using the AdamW optimizer with a $1e^{-4}$ weight decay and an initial learning rate of $5e^{-3}$, except for the VGG models that used an initial learning rate of $5e^{-4}$. The object detection models used an initial learning rate of $1e^{-3}$. The models were trained with a batch size of 64 over 10 epochs for classification and 200 epochs for object detection, using a cosine annealing learning rate scheduler that gradually decrease the learning rate towards 0. All convolutions were initialized using the Kaiming uniform method, and all batch normalization layers were initialized with a weight of 1 and a bias of 0. The Parametric LIF neurons all had an initial membrane time constant $\tau$ of 2, a membrane threshold of 1 and the ATan function as the surrogate function. Norm of the gradient values were clipped to a maximum of 1 to avoid exploding gradients. 
All presented results represent the best ones among 5 runs. 

All trainings were done with the SpikingJelly framework \cite{spikingjelly} using 16-bit automatic mixed precision, running on a 48-GB NVidia RTX A6000 and a 104-threads Intel Xeon Gold 6230R.

\subsubsection{Performance metrics}
For classification, we measure the performance by using an accuracy metric. For object detection, we report COCO mAP \cite{coco}, the mean Average Precision over 10 IoU ([.50:.05:.95]), as it is widely used for evaluating object detection models.

But assessing the performance of spiking neural networks is not limited to that, as multiple others features are needed to take advantage of their benefits when embedded in specialized hardware. For all our results we report the following metrics:
\begin{itemize}
    \item \textit{Number of parameters:} embedded systems have high constraints in term of memory, therefore it is important to design networks with a low number of parameters.
    \item \textit{ACCs:} spiking neural networks do not require multiplicative operations, enabling substantial energy savings on specialized hardware. Thus, we chose to report the number of operations of our SNNs by using the number of accumulations operations (ACCs), to accentuate the potential energy savings. Indeed, all spiking convolutions operations amount to ACCs, and each PLIF neuron only requires 1 ACC per timestep to update their potential. We did not count the ACCs in the batch normalization layers as they can be fused with the convolutional layers.
    \item \textit{Sparsity:} finally, we measured the number of spikes emitted after each activation layer to represent the global sparsity of the network compared to an fully dense equivalent DNN. Indeed, processing events with SNNs preserves the data sparsity. On specialized hardware, computations are only performed when there are spikes, therefore an highly sparse network would consume less power than its dense counterpart. The sparsity is obtained by averaging the number of spikes over the whole test set.
\end{itemize}

\subsection{Automotive Object Classification}

\begin{table}[]
\renewcommand{\arraystretch}{1.3}
\centering
\caption{Comparison with state-of-the-art models on Prophesee NCARS}
\begin{tabular}{lcccc}
\hline
\textbf{Methods}     & \textbf{Representation}     & \textbf{Network}   & \textbf{NCARS acc}\\ \hline
HATS \cite{ncars}       & TimeSurface   & N/A          & 0.902                        \\
Gabor-SNN \cite{ncars}   & Spike         & SNN       & 0.789                   \\
HybridSNN \cite{hybridsnn} & VoxelGrid  & SNN       & 0.77            \\
HybridSNN \cite{hybridsnn} & VoxelGrid  & SNN-CNN       & 0.906            \\
YOLE \cite{yole}       & VoxelGrid     & CNN          & 0.927                         \\
Asynet \cite{asynet}     & VoxelGrid     & CNN       & \textbf{0.944}                       \\
EvS-S \cite{graph} & Graph  & GNN       & 0.931                   \\
\hline
SqueezeNet 1.1 & VoxelCube & SNN       & 0.845                       \\ 
VGG-11 & VoxelCube & SNN       & \textbf{0.924}                       \\ 
MobileNet-64 & VoxelCube & SNN       & 0.917                       \\ 
DenseNet169-16 & VoxelCube & SNN       & 0.904                       \\ 
\hline
\end{tabular}
\label{tab:ncars}
\end{table}

For both classification datasets, the samples were resized to $64 \times 64$ pixels using nearest-neighbor interpolation to keep the input events binary.

We present the best accuracies obtained by our SNNs on the Prophesee NCARS dataset compared with other state-of-the-art models in Table~\ref{tab:ncars}. All of our models beat previous results for spiking neural networks and compete with the best neural networks in the literature. The spiking SqueezeNet, our smallest SNN, struggles to reach over 80\% test accuracy while spiking DenseNets, MobileNets and VGG are all capable to exceed 90\% test accuracy on NCARS.

\begin{table*}[]
\renewcommand{\arraystretch}{1.3}
\centering
\caption{Comparison between our spiking models on automotive classification}
\begin{tabular}{lcccccc}
\hline
\multicolumn{1}{c}{\multirow{2}{*}{\textbf{Models}}} & \multicolumn{1}{c}{\multirow{2}{*}{\textbf{\#Params}}} & \multicolumn{1}{c}{\multirow{2}{*}{\textbf{ACCs/ts}}} & \multicolumn{2}{c}{\textbf{NCARS}}                                   & \multicolumn{2}{c}{\textbf{GEN1 Classification}}        \\ 
\multicolumn{1}{c}{}                        & \multicolumn{1}{c}{}                          & \multicolumn{1}{c}{}                     & \multicolumn{1}{c}{\textbf{Accuracy $\uparrow$}} & \multicolumn{1}{c}{\textbf{Sparsity $\downarrow$}} & \multicolumn{1}{c}{\textbf{Accuracy $ \uparrow$}} & \multicolumn{1}{c}{\textbf{Sparsity $\downarrow$}} \\
\hline
SqueezeNet 1.0 & 0.74M & 0.05G        & 0.731 & 31.26\% & 0.627 & 6.65\%   \\
\textbf{SqueezeNet 1.1} & 0.72M & 0.02G        & \textbf{0.846} & 25.13\% & \textbf{0.674} & 6.79\%  \\
\hline
\textbf{VGG-11}     & 9.23M  & 0.61G        & \textbf{0.924} & 12.04\% & 0.969 & 14.69\%   \\ 
VGG-13     & 9.41M  & 0.92G        & 0.910 & 14.53\% & 0.970 & 19.03\%    \\ 
VGG-16     & 14.72M  & 1.26G       & 0.905 & 14.91\% & \textbf{0.977} & 18.79\%   \\ 
\hline
MobileNet-16  & 1.18M & 0.27G         & 0.842      &  17.57\% & 0.949 & 15.15\%        \\
MobileNet-32 & 7.41M & 1.06G         & 0.902   & 18.53\% & 0.955 & 14.37\%        \\
\textbf{MobileNet-64} & 18.81M & 4.20G         & \textbf{0.917}      & 17.14\%  & \textbf{0.966} & 30.60\%         \\
\hline
DenseNet121-16   & 1.76M   & 1.01G          & 0.889 & 27.99\% & 0.970 & 20.31\%   \\ 
DenseNet169-16   & 3.16M   & 1.19G          & 0.893 & 30.12\% & 0.969 & 23.12\%   \\ 
\textbf{DenseNet121-24}   & 3.93M   & 2.25G          & \textbf{0.904} & 33.59\% & \textbf{0.975} & 27.26\%   \\ 
DenseNet169-24   & 7.05M   & 2.66G          & 0.879 & 34.02\% & 0.962 & 28.29\%   \\ 
DenseNet121-32   & 6.95M   & 3.98G          & 0.898 & 38.32\% & 0.966 & 29.46\%   \\ 
DenseNet169-32   & 12.48M   & 4.72G         & 0.825 & 37.48\% & 0.967 & 40.35\%   \\ 

\hline
\end{tabular}
\label{tab:autoclassif}
\end{table*}

Table~\ref{tab:autoclassif} provides extensive results of all our spiking neural networks on both automotive classification datasets. Spiking SqueezeNet models, while having a very low number of parameters and number of ACCs per timestep, are not competitive with other architectures for the NCARS and the GEN1 classification datasets. Our spiking VGG models provide the best accuracies for both datasets, while maintaining a relatively low number of ACCs per timestep. But these architectures have an high number of parameters, making it difficult to embed them. Spiking MobileNets reach high accuracies but require high numbers of parameters and ACCs per timestep, penalized by the replacement of their depthwise seperable convolutions by normal convolutions. However, they are the only models for which the accuracy increases as the model gets bigger. Finally, spiking DenseNets reach competitive accuracies while requiring a low number of parameters and a moderate amount of ACCs per timestep. Using the densely connected layers of DenseNets, surrogate gradient method has no trouble to learn across 100+ spiking layers. The accuracy decreases however when the growth factor and the number of layers are both high, but we believe that better results could be achieved on these big networks with longer trainings. 

For SNNs to be truly efficient, they require both low sparsity and a low number of timesteps. All of our spiking neural networks have a sparsity inferior to 40\% of both datasets. As these SNNs operate on 5 timesteps, this means that they require at most twice the number of operations of an equivalent dense ANN. The operations would however consume less power on a specialized hardware, as they are simple ACCs. These sparsity results could further be improved by adding a regularization term to the loss, constraining the number of spikes emitted, as it was done in \cite{s2net}.

\subsection{Automotive Object Detection}

For object detection models, we used as backbones the best variants: VGG-11, MobileNet-64, DenseNet121-24. We evaluated our results on the Prophesee GEN1 Detection test set after having filtered boxes with diagonal smaller than 30 pixels as it is done in \cite{1mpx}. The spiking backbones were pre-trained on the NCARS dataset. Our results are presented in Table~\ref{tab:gen1}.

\begin{table}[]
\renewcommand{\arraystretch}{1.3}
\centering
\caption{Comparison with state-of-the-art models on Prophesee GEN1}
\begin{tabular}{lcccc}
\hline
\textbf{Methods}  &  \textbf{\#Params}  & \textbf{ACCs/ts}   & \textbf{Sparsity $\downarrow$} & \textbf{mAP $\uparrow$}   \\ \hline
Asynet \cite{asynet}      & 133M     & - & - & 0.15                       \\
MatrixLSTM \cite{matrixlstm}   & 65M    & - & -   & 0.31                   \\
RED \cite{1mpx} & 24M      & - & -     &   \textbf{0.40}          \\
\hline
VGG-11+SDD &  12.64M  & 11.07G & 22.22\%   & 0.174                       \\ 
MobileNet-64+SSD &  24.26M  & 4.34G & 29.44\%    & 0.147                       \\ 
DenseNet121-24+SSD  & 8.2M & 2.33G & 37.20\%     & \textbf{0.189}                       \\ 

\hline
\end{tabular}
\label{tab:gen1}
\end{table}

Our spiking models achieve competitive mAP with a small number of parameters and ACCs per timestep. We reach 0.19 COCO mAP with our DenseNet121-24 + SSD model, with only 8.2M parameters and 2.33G ACCs per timestep. Our spiking models outperform a traditional neural network with over 5 times more parameters. The three models show relatively similar performance, proving that spiking backbones are able to provide meaningful spike feature maps to do object detection on real-world event data. 

\section{Discussion}

\subsection{Influence of the number of timesteps and micro time bins\label{tbin}}

\begin{figure}[tb]
\centerline{\includegraphics[scale=0.42]{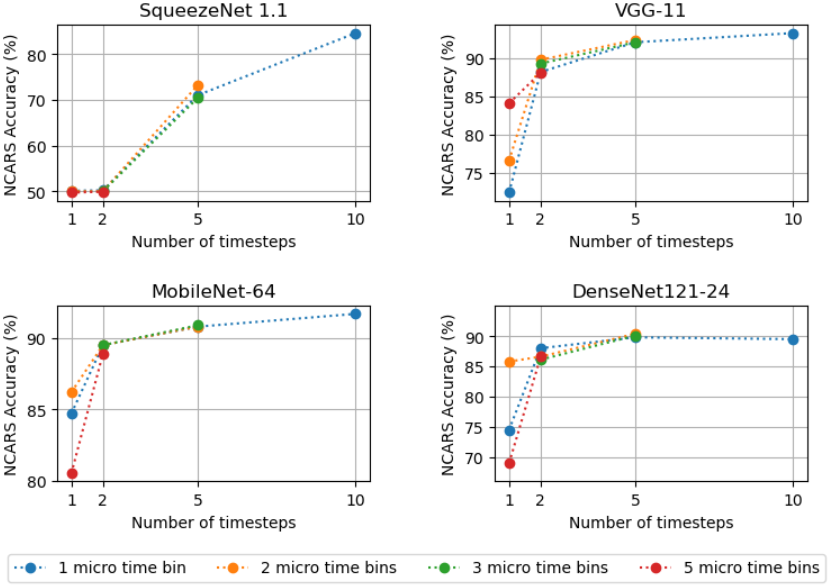}}
\caption{Influence of the number of timesteps and micro time bins on NCARS}
\label{fig:tbin}
\end{figure}

Spiking neural networks are recurrent neural network operating on a fixed number of timesteps $T$. Encoding event data in a representation that keep their temporal information can thus benefit SNNs. In section \ref{voxel}, we presented an event data encoding called voxel cubes, that preserve the temporal information of events while minimizing the number of timesteps. Indeed, the number of computations performed by SNNs increases linearly with the number of timesteps, it is thus important to keep it small.

Fig.~\ref{fig:tbin} plots the NCAR accuracy of our SNNs on selected combinations of number of timesteps and micro time bins. The number of timesteps remains the most important parameter for obtaining good accuracies with SNNs. Indeed, the best results are almost always achieved with 10 timesteps, the maximum value we tested. Results obtained with 1 timestep are significantly worse, even with a high number of micro time bins, proving if necessary that SNNs need to operate on several timesteps to be performant. Increasing the number of micro time bins do not always improve the results, even if it seems to help when the number of timesteps is low. We notice that our spiking SqueezeNet is unable to learn with 1 or 2 timesteps, indicating a strong dependence of this network on the temporality of the data.

Undoubtedly, these results depend on the samples duration and the data temporality. In our case, for our samples lasting 100ms, the best compromise between number of timesteps, number of micro time bins and accuracy seems to be 5 timesteps and 2 micro time bins, which is the encoding format we used for all models except SqueezeNet.

\subsection{Influence of Batch Normalization\label{bn} and PLIF neurons}

\begin{table}[]
\renewcommand{\arraystretch}{1.3}
\centering
\caption{Influence of Batch Normalization and PLIF neurons when training SNNs on NCARS.}
\begin{tabular}{lcccc}
\hline
\multicolumn{1}{c}{\multirow{2}{*}{\textbf{Models}}} & \multicolumn{4}{c}{\textbf{Accuracy $\uparrow$}}                                                                                   \\
\multicolumn{1}{c}{}                        & \multicolumn{1}{c}{\textbf{Normal}} & \multicolumn{1}{c}{\textbf{Post-BN}} & \multicolumn{1}{c}{\textbf{No BN}} & \multicolumn{1}{c}{\textbf{No PLIF}} \\
\hline
SqueezeNet 1.1 & 0.846 & 0.511 & 0.500 & 0.560                      \\ 
VGG-11 & 0.924 & 0.771 & 0.550 & 0.889                   \\ 
MobileNet-64 & 0.917 & 0.857 & 0.654 & 0.701  \\ 
DenseNet121-24 & 0.904  & 0.839 & 0.611  & 0.836     \\ 
\hline
\end{tabular}
\label{tab:bn}
\end{table}

Batch Normalization layers seem to be vital to the training of complex SNNs: as we can in Table~\ref{tab:bn}, removing them (while adding a bias term to the convolutions) makes our networks significantly less performant or they do not learn at all. More surprisingly, the placement of the batch normalization layers seems to also play a part in the efficiency of our networks, as placing batch normalization layers before the convolutions produce better results than placing them after. In DNNs, placing batch normalization layers before or after convolutions do not make a significant difference, as both placements provide benefits in training speed and convergence.

In our case, we believe that batch normalization layers placed before convolutions are effective in SNNs because they transform highly sparse feature maps of spikes into a dense decimal representation. As a result, the weights learned by convolutions are updated through backpropagation whether they have received spikes or not. Without batch normalization, only the weights receiving spikes would have been updated meaningfully, leading to slower convergence.

We restate that batch normalization can be used when training of SNNs because their parameters can be fused with the parameters of the subsequent convolutions. In light of this, we believe our findings on the placement of batch normalization layers in convolutional SNNs can make a difference in the training of larger and more complex SNNs.

On the other hand, PLIF neurons introduced in \cite{plif} also help during the training of our large spiking neural networks. Replacing them with simple LIF neurons (with a time constant $\tau$ of 2) leads to poorer accuracies for all networks.
Their presence is not as important as batch normalization layers but the SNNs seem to benefit from learning different time constants for each layer. It also reduces the number of hyperparameters to be tuned, so we can only encourage their use for the training of SNNs on event data.

\subsection{Influence of depthwise separable convolutions\label{dw}}

\begin{table}[]
\renewcommand{\arraystretch}{1.3}
\centering
\caption{Influence of depthwise separable convolutions when training spiking MobileNets on NCARS}
\begin{tabular}{lccc}
\hline
\textbf{Models}     & \textbf{Acc (dw sp conv)} & \textbf{Acc (normal conv)}  \\ \hline
MobileNet-16 & 0.842 & 0.906                       \\ 
MobileNet-32 & 0.902 & 0.898                      \\ 
MobileNet-64 & 0.917 & 0.807                \\ 
\hline
\end{tabular}
\label{tab:dwconv}
\end{table}

We used depthwise separable convolutions during the training of our spiking MobileNets for two reasons: the smaller number of parameters made the trainings faster and the larger networks were able attain better accuracies. As we can see in Table~\ref{tab:dwconv}, the 32 and 64 input channels variants of spiking MobileNets using depthwise separable convolutions reach higher accuracies than their normal convolution counterparts. 

We presume that the higher number of parameters induced by the normal convolutions makes the training of the spiking neural networks with surrogate gradient more difficult. As it is the case for the spiking VGGs, the accuracy actually drop when adding parameters to the normal convolutions MobileNets, while it increases for the MobileNets trained with depthwise seperable convolutions.

However, the smaller variant seems to benefit from the added parameters as the accuracy is 5\% higher with normal convolutions than with depthwise separable convolutions. We therefore recommend to use depthwise separable convolutions in SNNs only as a second resort, when the accuracy achieved with normal convolutions decreases as the networks get larger.

\section{Conclusion and future works}
We designed trained four different spiking neural networks models based on SqueezeNet, VGG, MobileNet and DenseNet, setting new state-of-the-art results on two automotive classification event datasets for spiking neural networks. We then used these networks combined to SSD bounding box regression heads to design the first spiking neural networks capable of doing object detection on the real-world event dataset Prophesee GEN1, achieving 0.19mAP with less than 10M parameters. All our SNNs are performant without requiring an high number of timesteps thanks to our \textit{voxel cube} event encoding. 

These results highlight the rapid progression of spiking neural networks in the last few years. For a long time restricted to small datasets, spiking neural networks now show their strengths when trained directly on temporal data. Future works would include the implementation of these spiking neural networks on a low-power neuromorphic hardware \cite{nassim}, \cite{loihi2}, which will enable power-efficient embedded applications.

\section*{Acknowledgment}

This material is based upon work supported by the French technological research agency (ANRT) through a CIFRE thesis in collaboration with Renault.


\end{document}